\documentclass[conference]{IEEEtran}
\IEEEoverridecommandlockouts
\usepackage{cite}
\usepackage{amsmath,amssymb,amsfonts}
\usepackage{algorithmic}
\usepackage{graphicx}
\usepackage{textcomp}
\usepackage{subcaption}
\usepackage{xcolor}
\usepackage{amsmath,amssymb}

\def\BibTeX{{\rm B\kern-.05em{\sc i\kern-.025em b}\kern-.08em
    T\kern-.1667em\lower.7ex\hbox{E}\kern-.125emX}}

\usepackage{amsmath,amssymb}
\DeclareMathOperator{\E}{\mathbb{E}}

\begin{document}

\title{Overcoming Missing and Incomplete Modalities with Generative Adversarial Networks for \\ Building Footprint Segmentation \\
}
\author{%
	\IEEEauthorblockN{Benjamin Bischke$^{1,2}$, Patrick Helber$^{1,2}$, Florian Koenig$^{1, 2}$, Damian Borth$^{2}$ and Andreas Dengel$^{1,2}$}\thanks{We gratefully acknowledge the support of NVIDIA Corporation with the donation of the DGX-1 used for this research. This work was partially funded by the BMBF Project DeFuseNN (01IW17002).
	}
	   \IEEEauthorblockA{$^{1}$TU Kaiserslautern, Germany}
	   \IEEEauthorblockA{$^{2}$German Research Center for Artificial Intelligence (DFKI), Germany}
		   \{Benjamin.Bischke, Patrick.Helber, Florian.Koenig, Damian.Borth, Andreas.Dengel\}@dfki.de
	 \\ 
}

\IEEEpubidadjcol
\IEEEoverridecommandlockouts


\IEEEpubid{\makebox[\columnwidth]{978-1-5386-7021-7/18/\$31.00 \copyright{}2018 IEEE~ \hfill} \hspace{\columnsep}\makebox[\columnwidth]{ }}


\IEEEoverridecommandlockouts
\IEEEpubid{978-1-5386-7021-7/18/\$31.00 ~\copyright~2018 IEEE}

\maketitle

\begin{abstract}
The integration of information acquired with different modalities, spatial resolution and spectral bands has shown to improve predictive accuracies. Data fusion is therefore one of the key challenges in remote sensing.
Most prior work focusing on multi-modal fusion, assumes that modalities are always available during inference. This assumption limits the applications of multi-modal models since in practice the data collection process is likely to generate data with missing, incomplete or corrupted modalities. In this paper, we show that Generative Adversarial Networks can be effectively used to overcome the problems that arise when modalities are missing or incomplete. Focusing on semantic segmentation of building footprints with missing modalities, our approach achieves an improvement of about 2\% on the Intersection over Union (IoU) against the same network that relies only on the available modality.
\end{abstract}


%
%
%
%
\section{Introduction}

Satellite data and aerial imaging is becoming more and more accessible in the recent years and changing the understanding of our planet. 
In order to make use of the valuable information that is contained in the sensory data and raw images captured by satellites, it is necessary to impose a layer that is able to reflect abstract geospatial features and real objects. 
With the success of deep learning, convolutional neural networks (CNNs) have shown to be a powerful approach to extract such higher level features and outperform many traditional computer vision methods \cite{lagrange2015benchmarking}. Applications of CNN's in this context are manifold and have been used to solve problems ranging from regression and classification \cite{he2018introduction} over to object detection to semantic \& instance segmentation \cite{bischke2017multi}.

One of the key challenges in remote sensing is the processing of multi-modal data sources and their fusion. Data fusion aims to integrate the information acquired with different spatial resolution, spectral bands and imaging modes from sensors mounted on satellites, aircraft and ground platforms to generated a combined representation that contains more detailed information than each of the individual sources \cite{kampffmeyer2017urban}. Fusing complementary information across modalities can lead to substantial synergy effects and improve the overall accuracy \cite{yokoya2018open, zhang2010multi}. Dedicated datasets and challenges such as the IEEE GRSS Data Fusion Contest \cite{datafusion} have been released recently in order to push research forward in areas of data fusion and multi-modal segmentation.

\begin{figure}[t!]
	\centering
	\includegraphics[width=0.7\linewidth]{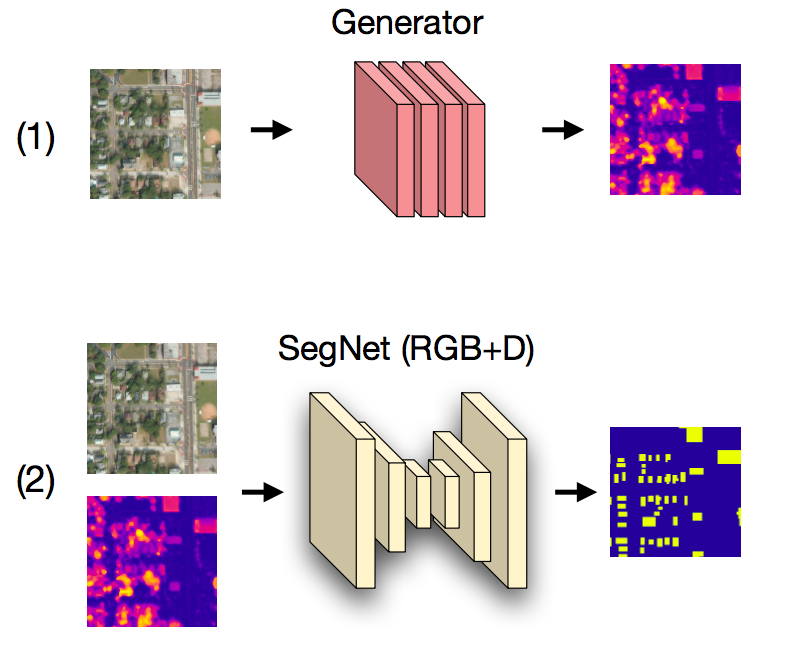}
	\caption{Convolutional neural networks are often trained with multi-modal information, such as the segmentation network shown in (2). Problems arise when modalities are corrupted, incomplete or missing and such networks can not be applied anymore. We use a generative neural network shown in (1) to overcome this problem by translating the input of the known modality (RGB) into a synthetic representation of the missing one (synthetic depth).}
	\label{fig:image}
\end{figure}
\IEEEpubidadjcol
 However, most prior work that focuses on multi-modal data fusion assumes that all modalities are available during inference and test time. This assumption can greatly limit applications of multi-modal analysis because in practice data collection process may likely generate data with missing modalities. Within the domain of remote sensing, this is the rule rather than the exception due to effects such as (1) blocking of spectral responses for optical sensors in the presence of clouds, (2) adverse constellations of non geostationary satellites at particular points in time and (3) corrupted/incomplete sensory data because unexpected exposures such as space debris and solar flares. In all of these scenarios, models that have been trained on multi-modal information can not be used anymore and a fall back strategy to a modal that only uses the available modality is often necessary.
 
 Within the scope of this work, we consider the problem of missing modalities on the exemplary use case of building footprints segmentation in a multi-modal setup. Our segmentation network, depicted in (2) in Fig \ref{fig:image}, is trained on two modalities relying on optical (RGB) and depth information. Our goal is to apply this model even in the absence of one modality by using a second CNN that is able to create a synthetic representation for the missing modality. Please note, that our approach is not restricted to high resolution data or to the absence of depth modality only. Our approach can generalize to other modalities as well, with the underlying assumption that the two modalities share a common feature space that allows the translation from one modality to another one.
 
 The contributions of this work are as follows:
\begin{itemize}
	\item We show that Generative Adversarial Networks (GANs) can be effectively used overcome the problems that arise when modalities are missing, incomplete or corrupted during inference time. Focusing on semantic segmentation of building footprints, our approach achieves an improvement of about 2\% on the Intersection over Union (IoU) of the building class against the same network that relies only on the available modality.
	\item To the best of the authors knowledge, this is the first work in literature which uses GANs not only for data augmentation during training but specifically solve the problem of incomplete and missing modalities during inference with satellite image data.\newline
\end{itemize}

%
%
%
%

\section{Related Work}
Our work is closely related to the following research areas:


\subsection{Multi-modal Segmentation}\label{AA}

Multi-modal data fusion for semantic segmentation is an active research field and different strategies ranging from early fusion to in-fusion and late fusion have been proposed. Audebert et. al \cite{audebert2017beyond} explore different fusion strategies within the context of multi-modal segmentation in remote sensing. In their experiments, fusion with the FuseNet architecture \cite{hazirbas2016fusenet} has proven to be slightly better compared to a late fusion approach. A similar idea has been explored previously by Huang. et. al \cite{huang2016building} in which a deconvolutional network with two parallel streams on RGB and NRG band combinations is used to fused the predictions of the two streams. Since we achieved with early fusion of channel stacking similar results compared to such a fusion approach, but with less convolutional filters and less computational load, we are fusing the modalities in this work via channel stacking.

\subsection{Overcoming Missing Modalities}\label{AA}
While there has been a lot of research in the field of reconstructing incomplete and sparse data via low rank recovery \cite{brand2002incremental} or factorization \cite{acar2011scalable}, there is not much work dealing with the complete absence of one modality during the inference. One of the first attempts of overcome this problem are \textit{Hallucination Networks} proposed by Hoffman et al.\cite{hoffman2016learning}. In this approach, CNNs are used to extract features for a particular modality and the predictions are obtained by a late fusion of all modalities. An additional network, the hallucination network, is then trained via regression in a way that features of the hallucination network extract similar features than the network which operates on the real modality. In case of a missing modality, the network trained on the real modality can be replaced by the corresponding hallucination network. Kampffmeyer et. al \cite{kampffmeyer2017urban} successfully applied this approach on satellite data to overcome missing depth information on land use and land cover segmentation.

\subsection{Generative Adversarial Networks}\label{AA}
The framework with GANs \cite{goodfellow2014generative} has achieved impressive results on image generation in the recent years. In the GAN setup, a discriminator and a generator play a zero sum game. Thereby, the generator tries to fool the discriminator, which on the other hand tries to distinguish between real and generated data samples. Due to the instability during training various improvements with respect to new training objectives \cite{arjovsky2017wasserstein} 
and combination with other models \cite{makhzani2015adversarial} have been proposed. In this work, we use GANs to capture the data distribution of modalities, such that in case of a missing modality we can generate the corresponding synthetic data and use it for inference.
\subsection{Image-to-Image Translation}\label{AA}
Image-to-image translation is the problem of translating one given representation of a scene from a source domain into the representation of another target domain. Please note, that source and target domain do not need to be necessarily different. 
One of the first unified framework which is also used in our work, was proposed by Isola et. al. \cite{isola2017image}. Their approach uses GANs in a conditional setting. Recent studies extend this work and learn the image translations in an unpaired manner without supervision. Since this represents an ill-posed problem, additional constraints are required to preserve certain properties in the image representation of the source domain. Various approaches have introduced additional constraints ranging from pixel values \cite{shrivastava2017learning} to cycle consistent training objectives \cite{zhu2017unpaired}. Liu et. al. \cite{liu2017unsupervised} proposed the UNIT framework, in which the image representations of two different domains are mapped and reconstructed through a shared latent space. 
Our work has close connections to such a multi-domain image translation, however we do not explicitly enforce additional constraints such as a shared latent space. 
Closest to our work are approaches in \cite{mou2018im2height, srivastava2017joint} which learn directly the mapping from RGB to synthetic Depth.








%
%
%
%
\section{Approach}
\begin{figure}
	\centering
	\includegraphics[width=0.99\linewidth]{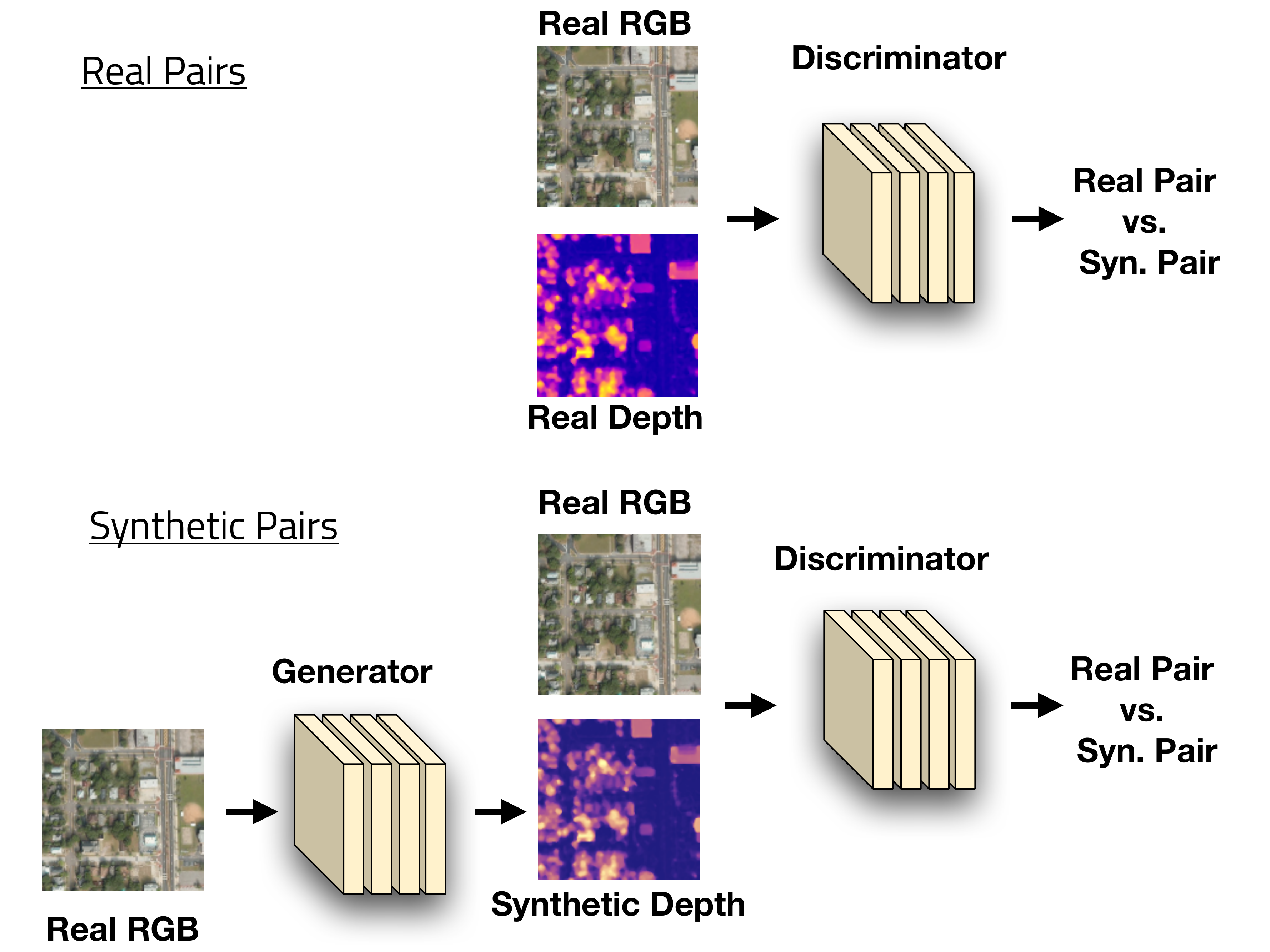}
	\caption{Setup of the Adversarial Training. We randomly sample from pairs of (1) RGB with real Depth and (2) RGB with synthetic Depth. Synthetic Depth images are generated by a Generator Network. }
	\label{fig:training}
\end{figure}

In this section, we describe our two step approach to overcome the effects of missing modalities with Generative Adversarial Networks. We first train a segmentation network on multi-modal input (RGB and Depth) for building footprint segmentation as depicted in Fig \ref{fig:image} (2). We then use a conditional GAN to capture the data distribution of the depth channel with respect to the RGB image. More formally, the conditional GAN is trained to learn the mapping from the RGB image $x$ and noise vector $z$ to a synthetic depth image $y$ with $G:{x, z} \to {y} $. 
The generator G receives the noise vector $z$ and RGB image $x$ as input and generates a synthetic depth image that looks indistinguishable from the real depth image. The discriminator D is trained adversarially to discriminate between the two pairs: (1) real pairs using RGB \& real depth and (2) on synthetic pairs using RGB \& synthetic depth. The adversarial training setup of the discriminator is depicted in Fig. \ref{fig:training}. After this training we only keep the generator G and use it to produce a synthetic depth image from RGB as input for segmentation network.

\subsection{Segmentation Network Architecture.}
The segmentation network used in this work is the fully convolutional network SegNet \cite{badrinarayanan2017segnet} which has an encoder-decoder architecture. The encoder has the same architecture as VGG16 \cite{simonyan2014very}, consists of 13 convolutional layers of 3x3 convolutions and five layers of 2x2 max pooling. The decoder is a mirrored version of the encoder which uses the pooling indices of the encoder to upsample the feature maps. 
SegNet offers a good trade-off between memory consumption and classification accuracy compared to other approaches such as U-Net and has been proven in the past to achieve good results on segmentation of building footprints \cite{bischke2017multi}. Please note, that we decided to use SegNet because of the good trade-off between memory and accuracy but it could be replaced in practice by any other segmentation network.

\subsection{Generative Adversarial Training}
The objective $L$ of our approach is formalized as follows:
\begin{equation}\label{eq:loss}
\begin{split}
L = \arg\min_{G}\max_{D} L_{GAN} (G, D) + \lambda * L_1(G).
\end{split}
\end{equation}
where $L_{GAN}$ represents the loss of the conditional GAN:
\begin{equation}
\begin{aligned}
L_{GAN}(G,D) &= \E_{x, y}[log D(x, y)]  + \\ 
&\quad \E_{x, z}[log(1 - D(x, G(x, z))]	\\
\end{aligned}
\end{equation}
Motivated by previous work, we add a further regularization term $L_1$ to the overall objective weighted by a factor $\lambda$.
\begin{equation}
\begin{split}
L_1 (G) = \E_{x, y, z}[||y - G(x, z)||_1] 
\end{split}
\end{equation}
This regularizer forces the generator to not only fool the discriminator, but also create synthetic depth images that look similar to the real depth maps in the L1-space. We use the L1 norm since it encourages less blurring compared to L2.

\subsection{Network Architectures}

We use similar architectures for the generator and discriminator network as presented in the work of Isola et. al \cite{isola2017image}. In the following, we discuss key points and differences.

\subsubsection{Generator Network}
The generator network in this work builds upon a CNN that uses an encoder decoder architecture. This architecture has been shown in the past to be successful to address pixel-based computer vision tasks ranging from semantic and instance-segmentation to image-to-image translation. Our encoder consists of 8 convolutions of size 4x4 with a stride of 2 followed by BatchNorm and ReLU layers. Each layer of the encoder network with the triple Convolution, Batchnorm and ReLU reduces consequently the spatial dimension of the input image. This yields to model that maps features from a high dimensional input space into a low dimensional latent space. The decoder has of the opposite structure than the encoder and learns the reverse mapping from latent space to a high dimensional output space.  
Since down-sampling of a high resolution input results in a loss of high frequency information, which we want to keep in the unknown output modality we add skip connections to each layer of the decoder. This follows the architecture of U-Net \cite{ronneberger2015u} where feature maps of the encoder are fused with the feature maps of the decoder using concatenation and an additional convolution. With these skip connections, high frequency information contained in the feature map of the encoder can be transferred to the up-sampled feature maps of the decoder which lack with detailed information to such an extend.


\subsubsection{Discriminator Network}

Our discriminator network is modeled as a CNN that tries to learn a binary classification between real versus synthetic pairs with respect to the two modalities of RGB and depth. 
Before we explain details of the network architecture, note that there is a regularization term L1 to the objective function $L$ in Eq. \ref{eq:loss}. The presence of this term forces the generator to capture low frequency information. However, generative models that only rely on L1 or L2 distances such as Variational Autoencoders \cite{kingma2013auto} have shown to fail to model high frequency information leading to generation of blurry images. 
This motivates to restrict the discriminator network to only capture the high frequency information. Isola et. al \cite{isola2017image} showed that is sufficient to restrict the attention of the discriminator to the structure in local image patches in order to model high frequency details. We use their proposed network architecture \textit{PatchGAN} that penalizes structure at a scale of patches. This discriminator is applied convolutationally across the image on smaller patches and the results are aggregated by averaging. Compared to discriminators that operate on the full image, this can be advantageous since with a smaller network fewer parameters are used, a faster runtime and processing of large images can be achieved. Because of the close connection of \textit{PatchGAN} to markov random fields \cite{li2016precomputed}, the network can be seen as a form of style loss.

%
%
%
%

\section{Experimental Results}


\subsection{Urban Mapper 3D Dataset}
We evaluate our approach on the Urban Mapper 3D Dataset. This dataset is comprised of 236 orthorectified color images, which have a size of 2048 x 2048 pixels. The satellite scenes cover the two cities Jacksonville and Tampa in Florida (USA) with a ground sample distance of 0.5 meters. In addition to optical RGB images, the dataset includes a digital surface model (DSM) and a digital terrain model (DTM) in the same spatial resolution for each tile. The surface and terrain models are represented as a single band of 32-bit signed floating point values that represent height (in meters) referenced to the WGS84 ellipsoid.
Ground-truth data is only provided for the public training set in form of instance and class labels for each building footprint. Since we do not have access to this private test set, we randomly split the training set with a ratio of 80/20 (for both locations) into new training and test sets. We subtract the DTM band from the DSM band in order to get the normalized height of each pixel. This representation in robust against topographical elevations such as hills and represents height information with respect to the surface as origin.
We use in the following experiments RGB images, the normalized depth and building class labels as ground truth. 

\subsection{Evaluation Metrics}
Following the evaluation protocol defined in \cite{maggiori2017can} for the segmentation of building footprints, we report our results on the same two metrics for all of the following experiments. The first one is the Intersection over Union (IoU) for the positive building class. This is the number of pixels labeled as building in the prediction and the ground truth, divided by the number of pixels labeled as pixel in the prediction or the ground truth. As second metric, we report accuracy, the percentage of correctly classified pixels.

\subsection{Experimental Setup -  Baselines \& Network Training}
In order to evaluate the performance of our approach, we first train different segmentation networks as a proxy to define our lower and upper bounds. We use RGB as known modality that is available during training and testing, whereas the depth modality is only available during the training phase. \newline

\begin{figure*}
	\captionsetup[subfigure]{justification=centering}
	\centering{
		\centerline{
			\includegraphics[width=0.145\linewidth]{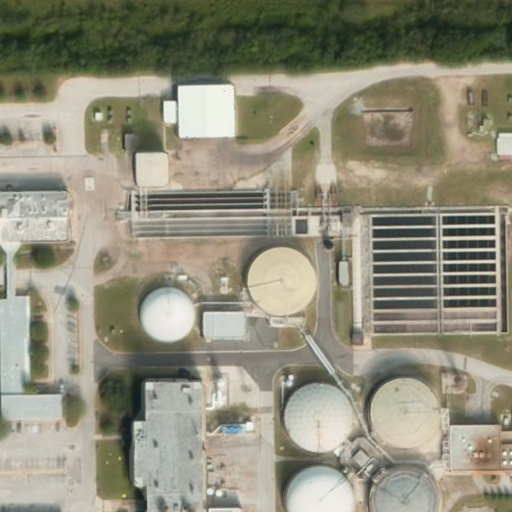}
			\includegraphics[width=0.145\linewidth]{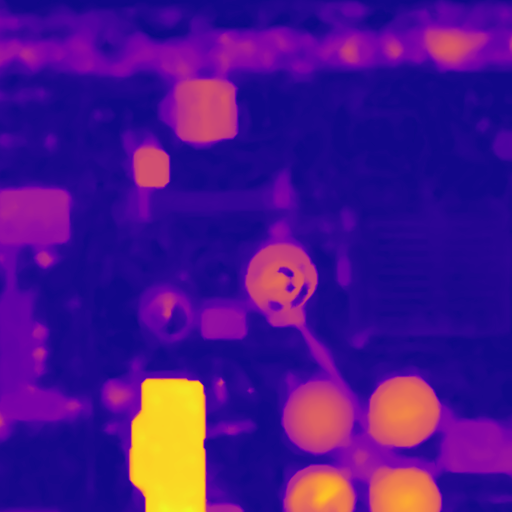}
			\includegraphics[width=0.145\linewidth]{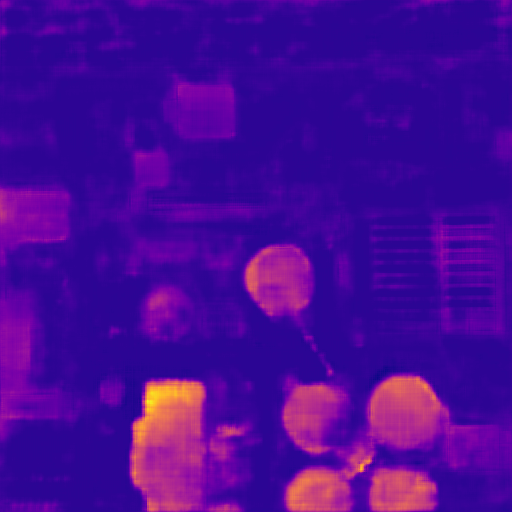}
			\includegraphics[width=0.145\linewidth]{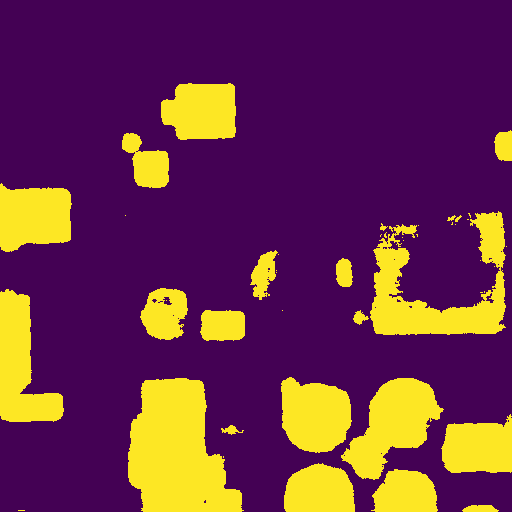}
			\includegraphics[width=0.145\linewidth]{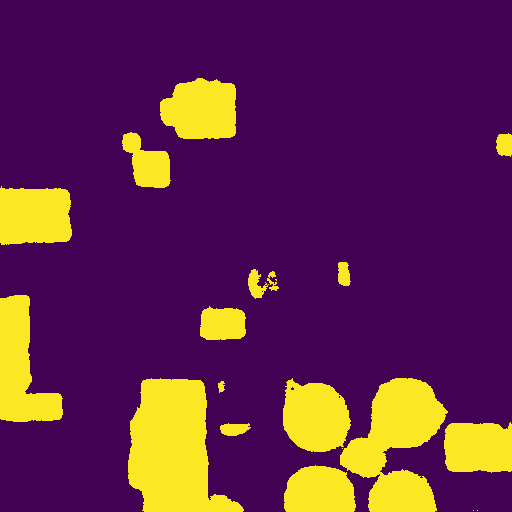}
			\includegraphics[width=0.145\linewidth]{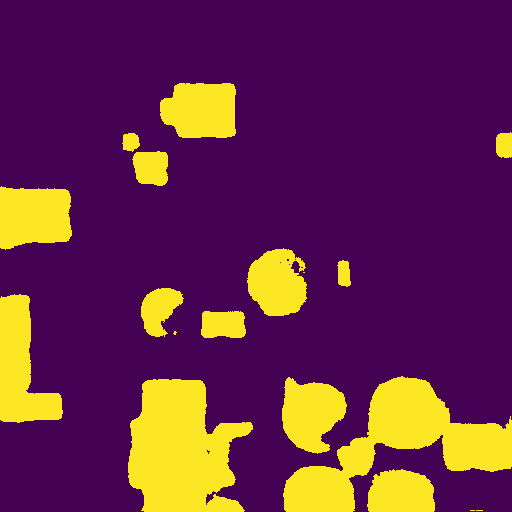}
			\includegraphics[width=0.145\linewidth]{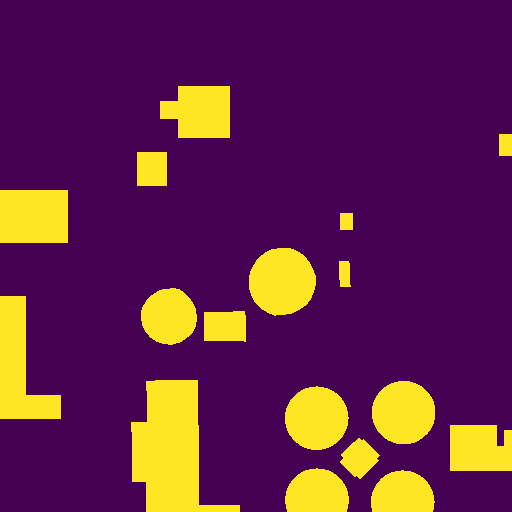}			
		}
		\medskip
		\vspace{-3pt}
		\centerline{
			\includegraphics[width=0.145\linewidth]{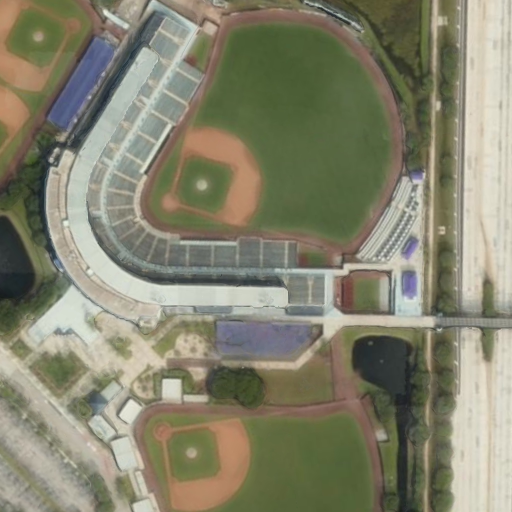}
			\includegraphics[width=0.145\linewidth]{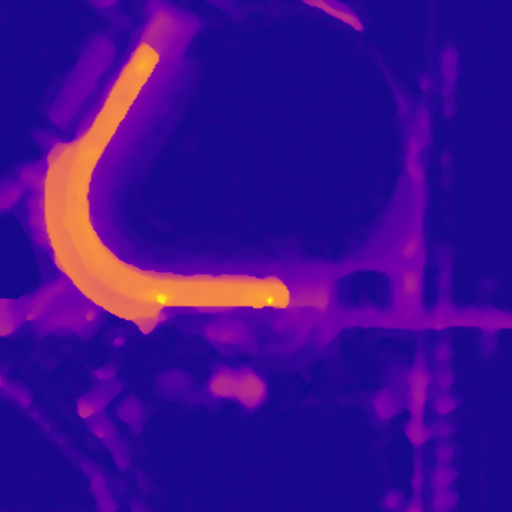}
			\includegraphics[width=0.145\linewidth]{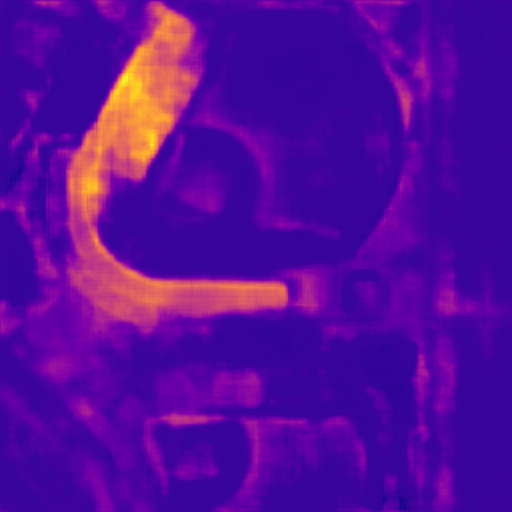}
			\includegraphics[width=0.145\linewidth]{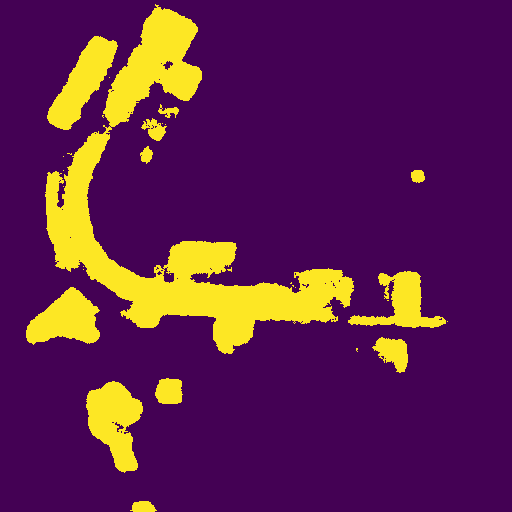}
			\includegraphics[width=0.145\linewidth]{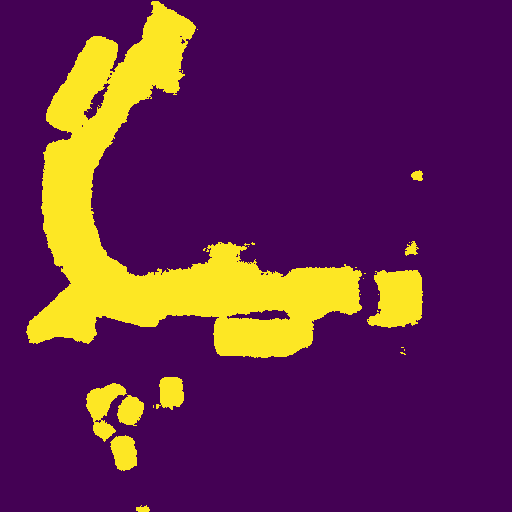}
			\includegraphics[width=0.145\linewidth]{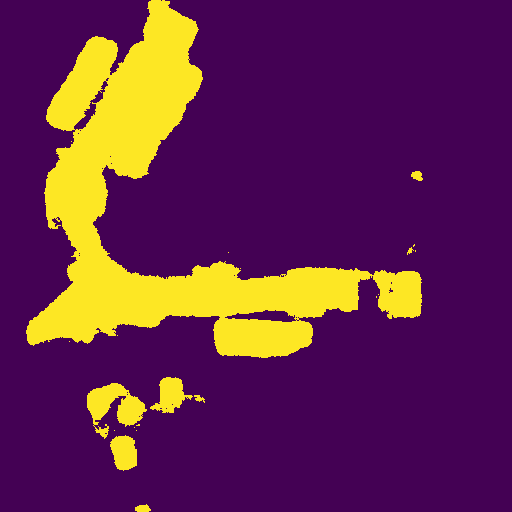}
			\includegraphics[width=0.145\linewidth]{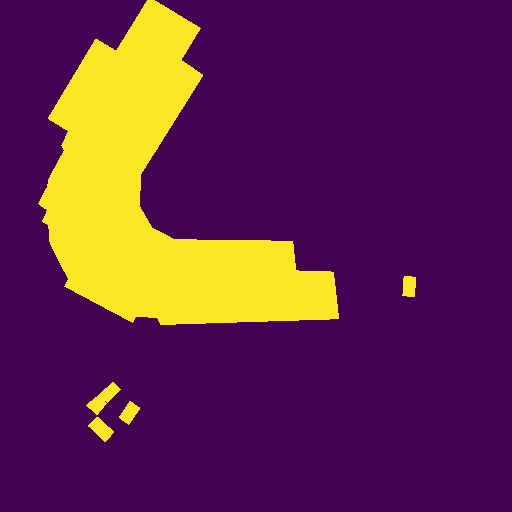}			
		}
		\medskip
		\vspace{-3pt}
		\centerline{
			\includegraphics[width=0.145\linewidth]{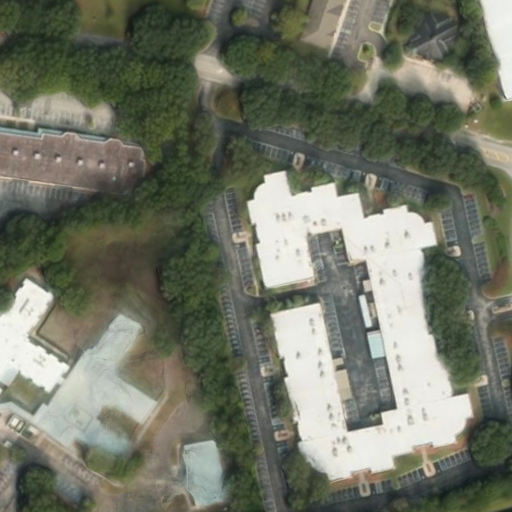}
			\includegraphics[width=0.145\linewidth]{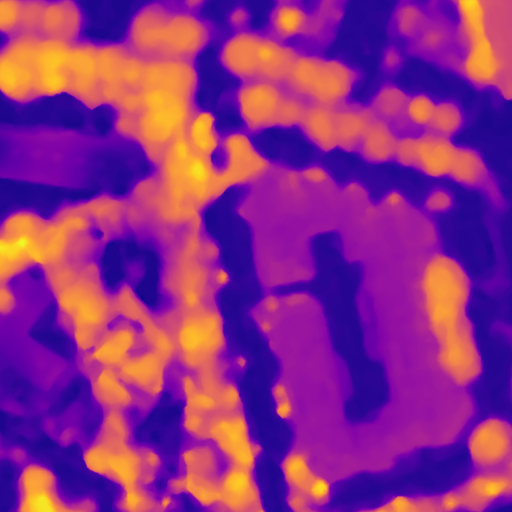}
			\includegraphics[width=0.145\linewidth]{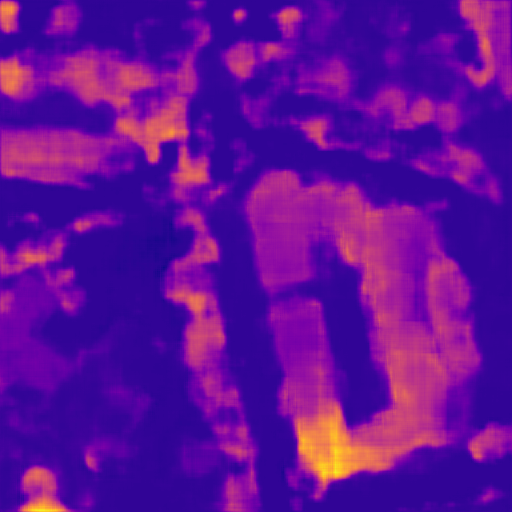}
			\includegraphics[width=0.145\linewidth]{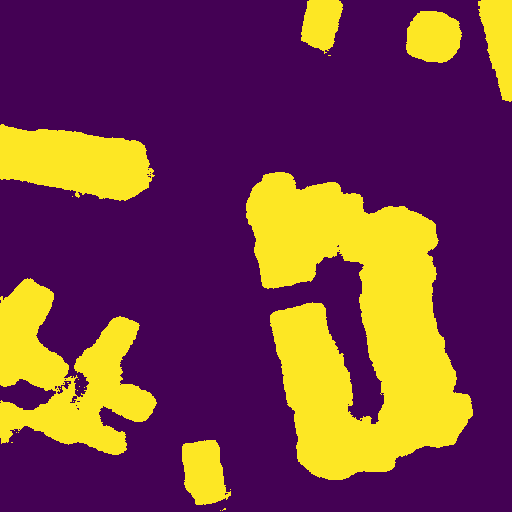}
			\includegraphics[width=0.145\linewidth]{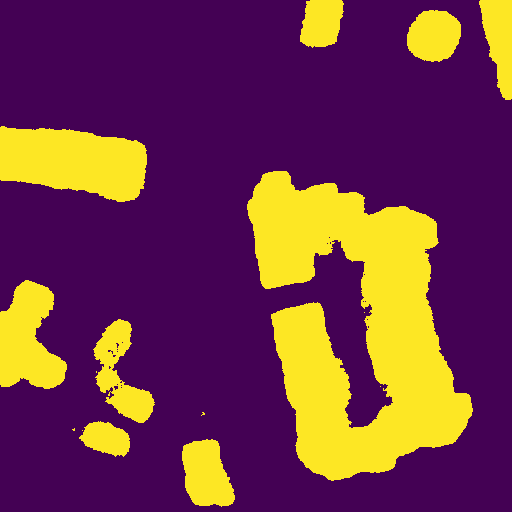}
			\includegraphics[width=0.145\linewidth]{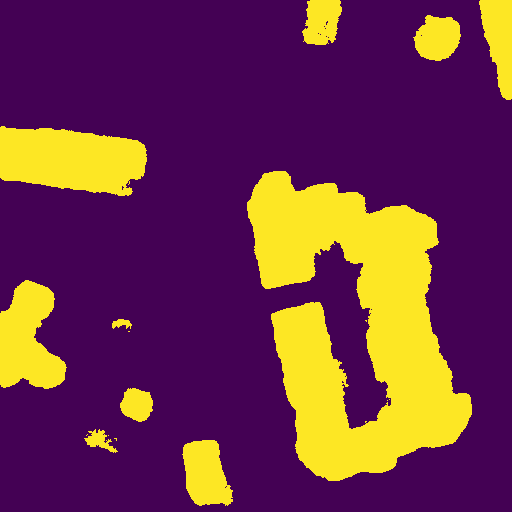}
			\includegraphics[width=0.145\linewidth]{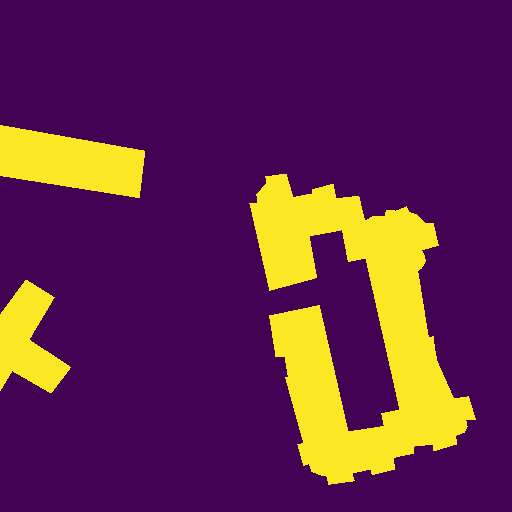}			
		}
		\medskip
		\vspace{-3pt}		
		\centerline{
		      \begin{subfigure}{0.145\linewidth}
		      	\includegraphics[width=\linewidth]{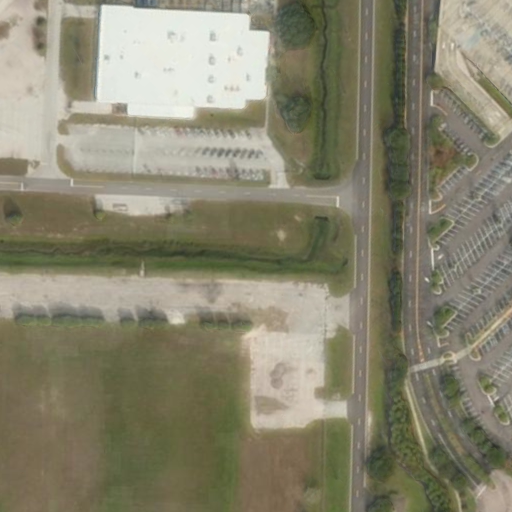}
		      	\caption{Real \\ RGB }
		      \end{subfigure}
		      \begin{subfigure}{0.145\linewidth}
		      	\includegraphics[width=\linewidth]{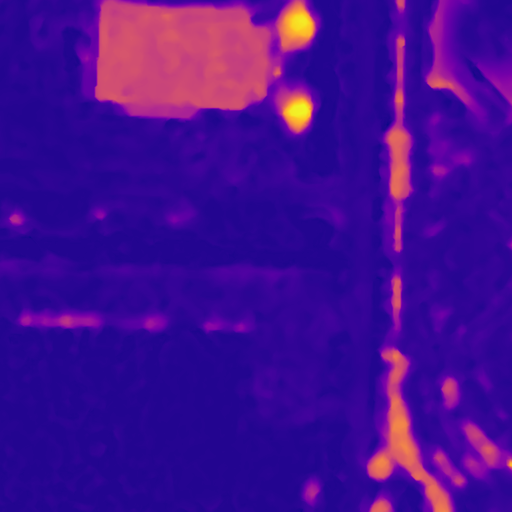}
		      	\caption{Real \\ Depth }
		      \end{subfigure}
		      \begin{subfigure}{0.145\linewidth}
		      	\includegraphics[width=\linewidth]{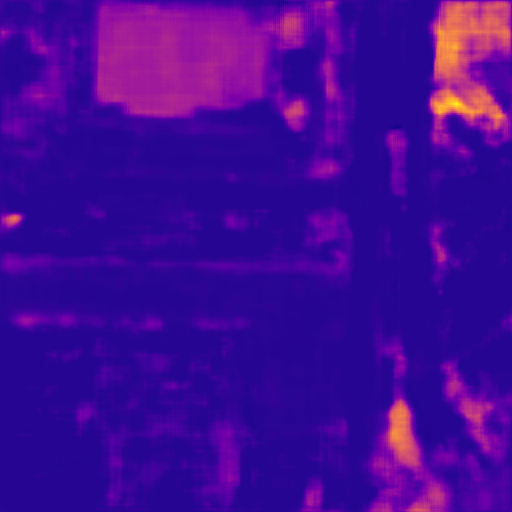}
		      	\caption{Synthetic \\ Depth }
		      \end{subfigure}
		      \begin{subfigure}{0.145\linewidth}
		      	\includegraphics[width=\linewidth]{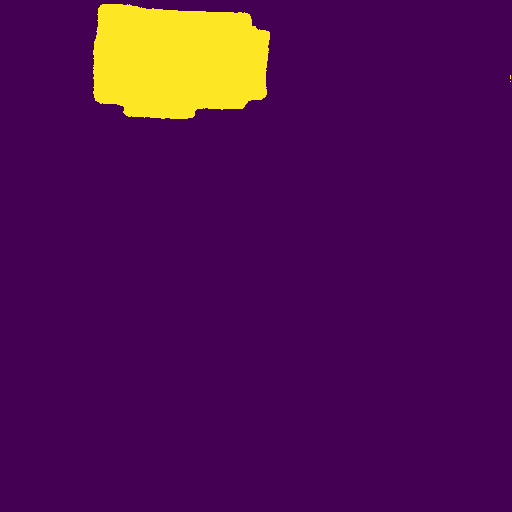}
		      	\caption{Prediction \\ (RGB only)}
		      \end{subfigure}
		      \begin{subfigure}{0.145\linewidth}
		      	\includegraphics[width=\linewidth]{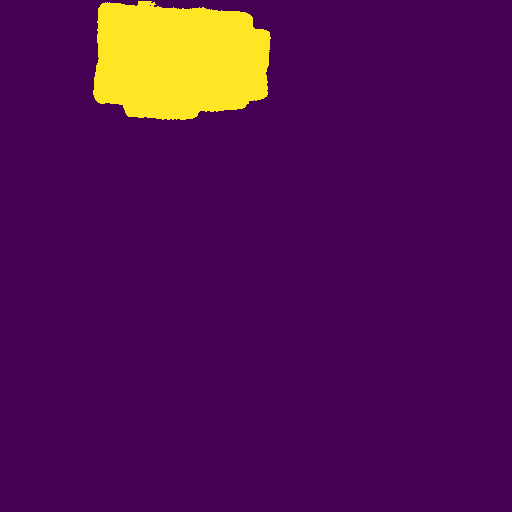}
		      	\caption{Prediction \\ (RGB \& Depth)}
		      \end{subfigure}
		      \begin{subfigure}{0.145\linewidth}
      		      	\includegraphics[width=\linewidth]{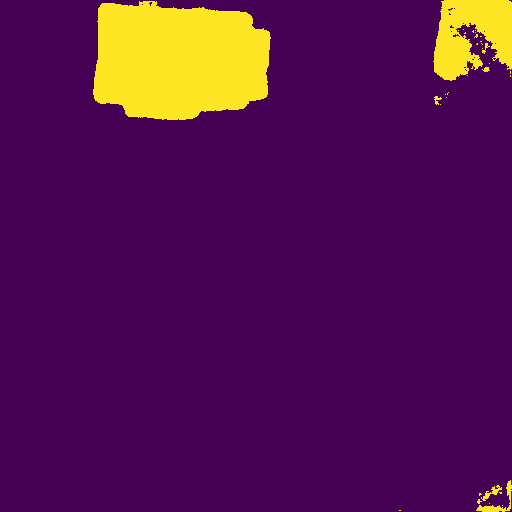}
      		      	\caption{Prediction (RGB \& Syn. Depth)}
		      \end{subfigure}
		      \begin{subfigure}{0.145\linewidth}
		      	\includegraphics[width=\linewidth]{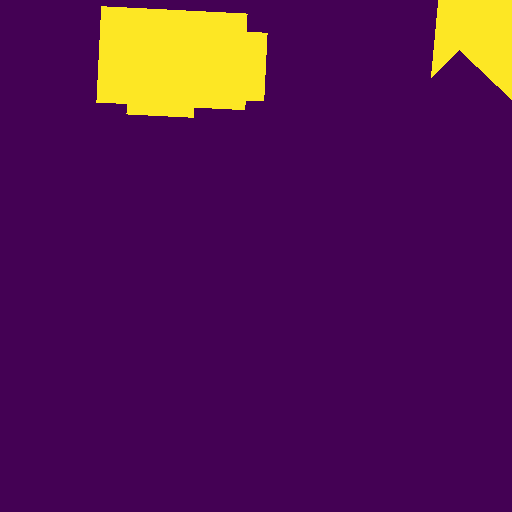}
		      	\caption{Ground \\ Truth }
		      \end{subfigure}
		}
	}
	\caption{Visualizations of real and synthetic modalities and the predictions of the corresponding modalities.}
	\label{fig:visual}
\end{figure*}
\begin{table}[t!]
	\centering
	\caption{Metrics of the Segmentation results for the different approaches. The numbers are stated as percentages}
	\label{tab:results}
	\small
	\begin{tabular}{cccc}
		\hline
		\textbf{Modality}                       & \textbf{\begin{tabular}[c]{@{}c@{}}IoU \\ Building\end{tabular}} & \textbf{\begin{tabular}[c]{@{}c@{}}IoU \\ Ground\end{tabular}} & \textbf{\begin{tabular}[c]{@{}c@{}}Pixel \\ Acc.\end{tabular}} \\ 
		\hline
		RGB only (Lower Bound)                  & 62.09                  & 92.91                        & 77.50                       \\
		RGB \& Partial Depth (Baseline)         & 62.34                     &   92.91                       & 77.63                       \\
		RGB \& Synthetic Depth & 63.96                    & 93.86                        & 78.91                       \\
		RGB \& Depth (Upper Bound)              & 65.58                     & 94.05                        & 79,82                       \\ \hline
	\end{tabular}
\end{table}
\subsubsection{Lower Bound - Training with RGB only}
We define the lower bound for this experiment by using a SegNet trained on RGB information only. Even though depth information is available during training, it is discarded. Any network trained on multi-modal information or missing modalities should be better than this network.
\subsubsection{Upper Bound - Training with RGB and Depth}
Our upper bound is determined by a SegNet which is trained and tested on all modalities. We use an early fusion approach with channel stacking to combine the three RGB channels and depth channel. It is worth mentioning, that there have been different approaches proposed in the past to fuse depth and optical RGB information. A very commonly used approach on satellite data is the FuseNet architecture 
\cite{hazirbas2016fusenet} in which two separate encoders fuse feature maps at each resolution via the \textit{add} operation. Since such an approach requires more parameters and we did not observe a significant improvement compared to an early fusion approach, we decided against it. Please note, that we want to show that the complementation of RGB with the depth modality using early fusion leads to better segmentation results compared to RGB only. Since RGB and depth information are available during both training and testing, this approach constitutes our upper bound.
\subsubsection{Baseline - Training with Incomplete Depth}
We use the same network architecture as above with an early fusion approach for our baseline. However, during test time only RGB information is available. In order to force the network to cope with the missing modality, we train the network with the following training procedure. We randomly (p=0.5) sample RGB and corresponding depth patches from two distributions: (1) one distribution with RGB and depth patches and (2) one distribution with RGB and no depth. The training setup ensures that if depth is missing, the network should still produce a meaningful output. Since there is on the other hand depth information available during training, this modality can be leveraged in order to extract features that produce better segmentation results than relying on RGB alone.
\subsubsection{Network Training}
We initialize the weights of the encoder where applicable, with the weights of a VGG16 \cite{simonyan2014very} model pre-trained on ImageNet. In case of more than three input channels in the first convolution, we initialize further channels with the average over the RGB channels. In the following, all networks are trained with SGD using a learning rate of 0.01, weight decay of 0.0005 and momentum of 0.9. We use the negative log likelihood loss and reduce the learning rate according to the poly policy \cite{zhao2016pyramid} (\textit{power=0.9}). Training is stopped after 30 epochs.
We extract 16 patches of size 512 x 512 pixels from each satellite image and use a batch size of 4 for the training. We apply randomly flipping in vertical and horizontal directions as data augmentation. 

\subsection{Synthetic Image Generation - Quantitative Results}
We train the GAN defined in section \ref{fig:training} for 200 epochs with a batch-size of 4 using the Adam optimizer. We start with a learning rate of 0.0002 and reduce it linearly to 0 starting from epoch 100. After this adversarial training the generator learned the mapping from RGB input to synthetic depth output.

The qualitative results when feeding the synthetic depth together with the RGB into the network trained on RGB and real depth can be seen in Table \ref{tab:results} (row 3). Our approach can not achieve the results as in the best case when all modalities are available during testing. However it is still significantly better compared to models that are relying on RGB only. The IoU of the positive building class is improved by about 2\% against the network that was trained on RGB only and 1.5\% better compared to our baseline that was explicitly trained on RGB and incomplete depth modality. Please note, that this improvement of 2\% is significant because of the large number of pixels that have to be classified. Similar results are achieved when considering the pixel accuracy metric. 

\subsection{Synthetic Image Generation - Qualitative Results}
We visualized in Fig. \ref{fig:visual}  for four different patches the RGB image (a), real depth (b) and synthetically generated depth image (c) as well as the segmentation results of the models using these different input modalities (d-g). When comparing the images of synthetic versus real depth (b \& c), a high similarity between the two representations can be observed. This includes sharp edges on building outlines, on shadows of buildings and accounts even for more complex building structures as shown in the first two rows. When comparing the predictions based on RGB only against RGB with synthetic depth (d \& e), an improvement in the segmentation mask can be observed when using synthetic depth. This accounts for large buildings as seen in the upper right part of row 1 and for small buildings shown in the lower left part of row 2.

In the third row of Fig. \ref{fig:visual} an interesting case of incomplete or corrupted data can be observed. The real depth image shows two vertical rows of trees (in orange-yellowish color), but in the optical image only one row of trees is present. It could be that case that the images of the two modalities were taken at two different timestamps. When this noisy multi-modal information is passed to the segmentation network, the parking space in the optical image is misleadingly classified as building because of the similar texture in the optical image and height cues in the depth image. The generator on the other hand did not predict any height cues at this position and consequently the parking space is correctly classified as ground.

The opposite effect can be observed in the fourth row. Here, information is lost near the image border in the upper right corner of the real depth image whereas the synthetic image contains a stronger signal of height. Consequently the upper right building is detected when relying on synthetic data but not when using the noisy but real modalities.
\section{Conclusion}
In this paper we addressed the problem of missing modalities during inference time in a setup in which models trained on multi-modal information. We showed that GANs provide a powerful approach to overcome this problems and better segmentation accuracies of about 2\% can be obtained compared to relying on the single, available modality only. 
It can be seen in Fig. \ref{fig:visual} that our model learned to estimate height not only for the building class which this work was focusing on primarily, but also other land cover classes such as trees. It is interesting to extend our segmentation approach also to additional land cover classes.
We are currently extending this work with other training objectives and generative models such as Structure to Signal Autoencoders \cite{folz2018adversarial}.


\bibliographystyle{IEEEbib}
\bibliography{missing_gan.bib}
\end{document}